# МОДЕЛИРОВАНИЕ ЭМОЦИЙ И ЧУВСТВ: КОЛИЧЕСТВЕННАЯ НЕЙРОСЕТЕВАЯ МОДЕЛЬ ЧУВСТВА ЗНАНИЯ

П.М. Гопыч

*Харьковский национальный университет им. В.Н. Каразина,
пл. Свободы 4, Харьков, 61077, Украина, pmg@kharkov.com*

*Анотація.* Вперше запропоновано кількісну нейросітьову модель емоцій та відчуттів, яка базується на існуючих даних про їх нейро- та еволюційно-біологічну природу, а також на нейросітьовій моделі людської пам'яті, що дозволяє окремо розглядати свідомі та підсвідомі процеси у їх залежності від часу. Як приклад запропоновану модель застосовано для кількісного аналізу відчуття знання.

*Abstract.* The first quantitative neural network model of feelings and emotions is proposed on the base of available data on their neuroscience and evolutionary biology nature, and on a neural network human memory model which admits distinct description of conscious and unconscious mental processes in a time dependent manner. As an example, proposed model is applied to quantitative description of the feeling of knowing.

**1. Введение**

Сознание стали рассматривать как основу человеческого поведения только с середины 20-го века, когда в результате когнитивной революции на смену господствовавшему до этого бихевиоризму в (западную) психологию пришел когнитивизм, ставший также основой исследований в области классического искусственного интеллекта, стремящегося смоделировать сознание средствами формальной логики в вычислительной среде. После неудачи таких попыток центр тяжести исследований сместился на нейронные науки и эволюционную биологию и, таким образом, возродился своеобразный «натурализм», когда психологи стали детально исследовать механизмы реализации феномена сознания у реальных людей – представителей вида *Homo sapiens*. При этом неожиданно выяснилось, что эмоции человека являются совершенно необходимой частью процесса принятия им решений. Так, например, пациенты с повреждением орбитофронтальной коры (части мозга, имеющей дело с социальными эмоциями) не могут принимать простейшие решения и часто действуют иррационально [1]. Под влиянием таких результатов возникает впечатление, что в основе сознания скорее эмоции чем процессы более высокого уровня (типа языка) и что эмоции играют решающую роль в различении сознательных и подсознательных процессов. Такая точка зрения не является общепринятой, но она коррелирует с новейшими гипотезами о критичности для сознания процессов в системе «мозг – тело – окружающая среда» и свидетельствует о важности выявления истинного значения эмоций для разрешения проблемы сознания и осознанного поведения человека.

К настоящему времени при исследовании эмоциональных состояний животных и человека в опытах с применением, в частности, функциональной магнито-резонансной и позитронно-эмиссионной томографии получен огромный фактический материал, который еще только предстоит обобщить в рамках единой интегрированной концептуальной модели эмоций. А пока даже создание концептуальных (и компьютерных) моделей хотя бы отдельных проявлений эмоций, например эмоциональной памяти, рассматривается как актуальная задача только для будущих исследований.

В настоящей работе на основе нейросетевой модели памяти [2], которая предоставляет, в частности, возможность раздельного описания сознательных и подсознательных процессов в их зависимости от времени, впервые предложена концептуальная модель для описания эмоций и чувств и ее количественная компьютерная реализация на примере чувства знания.

## 2. Чувства как прямое следствие эмоций

В литературе при обсуждении эмоций и чувств имеются многочисленные разночтения. Далее мы будем следовать подходу и определениям работы [1], которые в сжатой форме приведены ниже.

*Эмоции* – это специфически организованный набор химических и нейронных откликов, которые продуцируются мозгом при обнаружении им *эмоционального стимула* (например, объекта или ситуации), причем обработка этого стимула не обязательно является осознанной. *Эмоциональный отклик* – это специфическая реакция мозга на определенные классы объектов или событий, возникшая в результате эволюции и характеризующаяся определенным (стереотипным) набором действий. Главный *объект эмоционального отклика* – это тело, его внутренняя среда, внутренние органы, скелето-мышечная система, а также сам мозг. Результатом отклика, направленного на тело, есть *эмоциональное состояние*, включающее соответствующий гомеостатический баланс, специфическое поведение и выражение лица. Результатом отклика, направленного на мозг (через тело), может быть, например, изменение внимания к эмоциональному стимулу. Большинство эмоциональных откликов наблюдаются либо невооруженным глазом, либо в результате психофизиологических или нейрофизиологических измерений, а также в результате эндокринных анализов. Таким образом, эмоции не являются субъективными, личностными или неуловимыми. Они «публичны» и их нейробиология может объективно изучаться (в частности, регистрироваться «детектором лжи») и у человека, и у других живых существ.

*Чувства* – это ментальное представление физиологических изменений вызванных эмоциями и, следовательно, они являются их прямым следствием. В отличие от эмоций чувства действительно субъективны и личностны, хотя и доступны объективному научному анализу с помощью методов, обычных при изучении других когнитивных явлений. Чувство соответствующей эмоции обеспечивает организму ментальную готовность, настороженность к определенным эмоциональным ситуациям, обостряет их ощущение, улучшает обучаемость и увеличивает вероятность того, что в будущем похожие ситуации будут успешно преодолены.

При исследовании животных и человека идентифицированы нейронные системы, вовлеченные в продуцирование эмоций. Такие структуры мозга как миндалины или вентромедиальная префронтальная кора их «запускают», а исполнителями эмоций являются структуры в гипоталамусе, базальном переднем мозге и в основании мозга. Они непосредственно сигнализируют (химически или неврологически) тем частям тела и мозга, изменения в которых создают эмоциональные состояния (разным эмоциям могут соответствовать разные распределения нейронной мозговой активности).

## 3. Концептуальная модель эмоций и чувств, учет настроений

На основе известных данных о системах и процессах зарождения и исполнения эмоций и чувств предлагаемая концептуальная модель явно выделяет место и роль в них неосознаваемой (импликативной) и осознаваемой (экспликативной или декларативной) памятей, учитывает наличие автономной гормонально-стрессовой системы и выходов на автономную и/или контролируемую сознанием сенсорно-моторную систему.

Предлагаемая модель предполагает (рис. 1), что исходный внешний или внутренний стимул (1) поступает на вход долговременной импликативной памяти (2). Если он распознается как один из эмоциональных (стрессовых) стимулов, то немедленно продуцируется автономный гормонально-стрессовый отклик (3), посредством которого вызывается стереотипная поведенческая реакция (4), специфическая для каждого эмоционального стимула и реализуемая автономной сенсорно-моторной системой. Эта чисто *эмоцио* *нальная* поведенческая реакция возникает по схеме 1-2-3-4 без участия декларативной памяти, т.е. бессознательно и, следовательно, быстро.

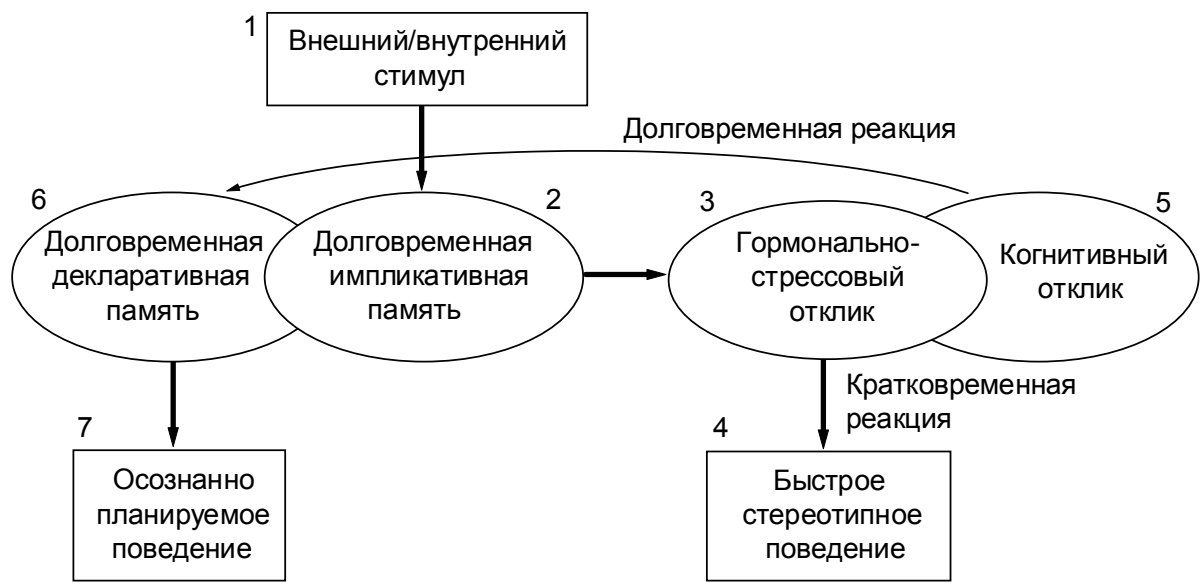

Рис. 1. Схематическое представление концептуальной модели возникновения быстрой стереотипной, контролируемой преимущественно эмоциями (цепочка 1-2-3-4), и более медленной осознанной, контролируемой преимущественно чувствами (цепочка 1-2-3-5-6-7), поведенческой реакции на эмоциональный (стрессовый) стимул.

Кроме отклика, направленного на тело (блоки 3 и 4), имеет место и когнитивный отклик (5), направленный на мозг. Посредством когнитивного отклика после распознавания эмоционального стимула мобилизуется дополнительное внимание к нему и оказывается влияние на декларативную (6) и импликативную (2) памяти о стимуле в ходе уже более медленного процесса его закрепления и/или обновления, происходящего при участии сознания. Именно таким путем (цепочка 1-2-3-5-6) в долговременной памяти (2 и 6) формируется ментальное представление о вызываемых эмоцией физиологических изменениях, т.е. *чувство* о ней. Таким образом, выброс гормонов (блок 3) оказывает влияние и на имеющуюся (врожденную или приобретенную) память об эмоциональных стимулах, и на возникновение новой памяти о ранее незнакомых таких стимулах. В будущем (или сразу) *чувство* используется организмом для осознанного планирования бесконечно разнообразного, хотя и относительно медленного, поведения (7) в ответ не только на уже предъявленный, но и на ожидаемый («предчувствуемый») эмоциональный стимул, что повышает приспособляемость организма (см. раздел 2).

Из сказанного следует, что предлагаемая модель в ответ на эмоциональный стимул допускает чисто *эмоциональную* (без участия сознания) быструю поведенческую реакцию (цепочка 1-2-3-4) и контролируемую сознанием более медленную *чувственную* поведенческую реакцию (цепочка 1-2-3-5-6-7), которая всегда протекает при наличии *эмоционального фона* (или *настроения*), интенсивность и характер которого определяется интенсивностью и характером влияния на цепочку 1-2-3-5-6-7 ее гормонального звена 3. Если гормонально-стрессовая система не оказывает влияния на поведенческую реакцию в ответ на эмоциональный стимул (цепочка 1-2-6-7), то это, вероятнее всего, паталогия или аномалия. Наиболее типичны промежуточные ситуации, когда поведенческие реакции или преимущественно чувственные (в значительной мере контролируются сознанием), или преимущественно эмоциональные (в значительной мере бессознательные и автоматические).

## 4. Численная реализация модели

Превратим предложенную концептуальную модель (раздел 3) в количественную, применив для описания долговременных импликативной и эксплиативной памятей (блоки 2, 6 на рис. 1) имеющуюся нейросетевую модель памяти [2], допускающую их раздельное описание и схематически представленную на рис. 2. Гормонально-стрессовый отклик (3) учтем путем введения числового параметра, характеризующего интенсивность эмоционального фона, а количественная модель когнитивного отклика (5) заложена, полагаем, главным образом в алгоритм обучения нейросетевой ячейки памяти (обсуждение этих аспектов модели выходит за рамки настоящей работы).

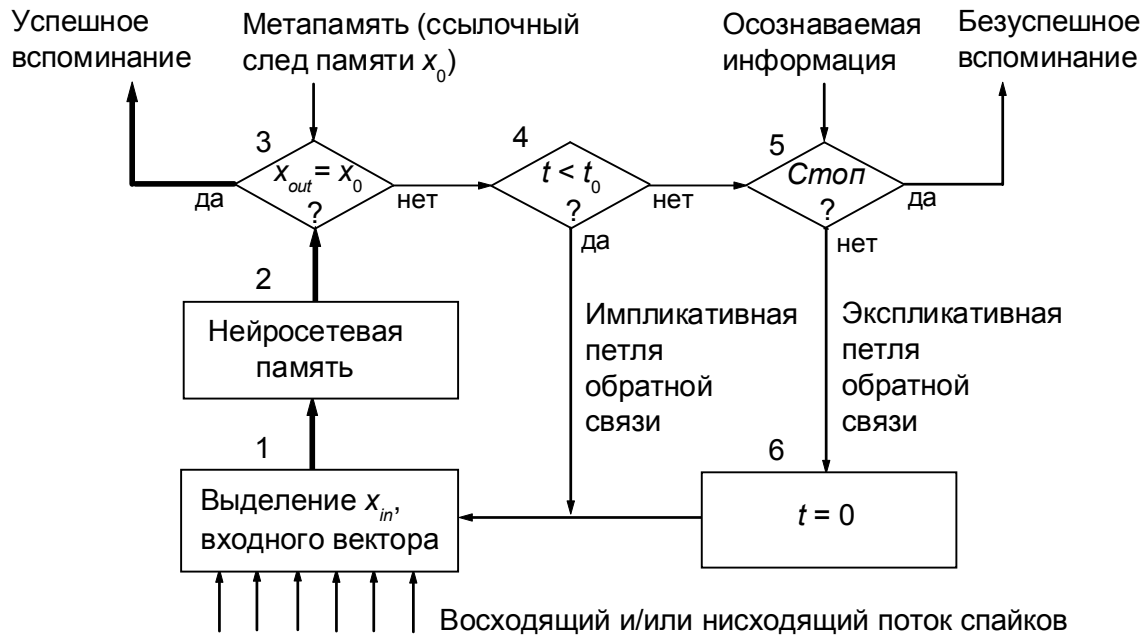

Рис. 2. Модельное соотношение между импликативной памятью, эксплиативной памятью и метапамятью. Жирные стрелки обозначают путь распространения синхронизованных пакетов нервных импульсов или «спайков» (векторов $x_{in}$, $x_{out}$ или $x_0$) [2].

Модель памяти [2] предполагает (рис. 2), что запоминаемая информация $x_0$ хранится в нейросетевой ячейке памяти (2), которая является частью изображенной на рисунке открытой системы, и посредством входных ($x_{in}$ формируемых блоком 1) и выходных ($x_{out}$, формируемых ячейкой памяти блок 2) векторов может быть связана со своим ссылочным следом в метапамяти (тоже $x_0$) и с внешней эксплиативной информацией. Локализация ячейки (2) со всеми ее связями (рис. 2) происходит импликативно. Петля 1-2-3-4-1, повторяемая с частотой $f$, моделирует импликативное вспоминание (параметр $t_0$ – его характерное время), петля 1-2-3-4-5-6-1 – эксплиативное вспоминание, с использованием (5) внешней осознаваемой информации. Вероятностные характеристики вспоминания следа $x_0$ количественно определяются свойствами ячейки (2) и вычисляются по правилам из работы [2], а долговременная стратегия вспоминания зависит от внешней (5) осознаваемой информации. $x_0$ на рис. 2 – это след памяти об одном из эмоциональных стимулов, содержащий всю информацию о необходимых в ответ на данный стимул гормональных, поведенческих и когнитивных откликах.

## 5. Модельный анализ чувства знания

В лабораторных условиях чувство знания (ЧЗ) исследуют обычно на примере явления «на кончике языка» (НКЯ), когда человек в данный момент не может вспомнить нуж ное слово, но обладает его ЧЗ (уверен, что его знает). Яркий пример проявления затяжного состояния НКЯ и ЧЗ описан в известном рассказе А.П. Чехова «Лошадиная фамилия», а их количественный нейросетевой анализ – в работе [3] (там же см. дополнительную литературу).

Источником ЧЗ является факт идентификации в метапамяти (памяти о памяти) ссылочного следа $x_0$ (см. рис. 2). Когда речь плавная, ЧЗ каждого произносимого слова не успевает осознаваться, так как вслед за неосознаваемой идентификацией следа $x_0$ следует его неосознаваемое извлечение и воспроизведение из памяти (2), затем идентификация новой ячейки памяти и соответствующего ей в метапамяти следа $x_0$ для следующего слова и т.д. Но в случае НКЯ след $x_0$ за время $t_0$ по петле 1-2-3-4-1 импликативно не извлекается и к процессу вспоминания подключается петля 1-2-3-4-5-6-1, предполагающая обмен (5) осознаваемой информацией с внешней для ячейки памяти (2) средой. В данный момент и происходит осознание возникшего ЧЗ, решающим образом влияющего на выбор дальнейшей стратегии вспоминания, которая осознанно корректируется всякий раз, когда происходит обращение к блоку 5. В описанной А.П. Чеховым ситуации этот процесс продолжается около суток пока новая подсказка не приводит к немедленному, неожиданному и долгожданному вспоминанию, которое сопровождается сначала быстрой, преимущественно стереотипной, автоматической и бессознательной (цепочка 1-2-3-4 на рис. 1), а затем долговременной и преимущественно осознанно планируемой (цепочка 1-2-3-5-6-7) поведенческой реакцией. При этом интенсивность эмоционального фона вначале бо́льшая, а затем ме́ньшая. Отметим, что следы $x_0$ памяти о вспоминаемом слове и об эмоциональном отклике на его вспоминание хранятся в разных, но однотипных ячейках памяти, вероятностные характеристики которых количественно описываются одинаковыми формулами [2]. Численные примеры приведены в [3].

## 6. Заключение

Подход [1] и реализующая его модель, предложенная в настоящей работе, допускают описание, вообще говоря, произвольных эмоциональных проявлений и у человека, и у животных (что важно для выбора этики наших взаимоотношений с ними). В частности, становится возможной количественная модельная интерпретация таких общеизвестных следствий из опыта поколений как «утро вечера мудренее» или «не сдерживай порывов, идущих от души», а также феноменов возникновения чувства страха только после завершения вызвавшего его скоротечного эмоционального стимула или вспоминания после пережитого стрессового состояния следов почему-то утраченной ранее части декларативной памяти. Однако наиболее важным представляется то, что модель явно учитывает и допускает количественное описание соотношений между эмоциями, чувствами, настроениями, связанными с ними поведенческими реакциями, сознанием и подсознанием (включая феномен зомби). Например, согласно модели, эмоции не разделяют сознательные и подсознательные процессы, а только создают для них эмоциональный фон.

## Литература

1. Damasio A.R. The feeling of what happens: body and emotion in the making consciousness. New York, Harcourt Brace, 2000. – 396S.
2. Гопыч П.М. Определение характеристик памяти. // Краткие сообщения ОИЯИ. – 1999. – № 4[96]-99. – С.61-68. См. также http://dbserv.jinr.ru/text/JinrRC/RC4-99/text7.html.
3. Гопыч П.М. Трехэтапная количественная нейросетевая модель явления «на кончике языка». // Труды IX-й Международной конференции «Знание-диалог-решение» (KDS-2001). – СПб, Россия, 2001. – С.158-165. См. также http://arXiv.org/abs/cs.CL/0107012.